\documentclass[11pt,a4paper]{article}

\usepackage[hyperref]{acl2021}
\usepackage{times}
\usepackage{latexsym}

\usepackage{microtype}

\aclfinalcopy 
\def\aclpaperid{163} 

\usepackage{amsmath}
\usepackage{amssymb}
\usepackage{booktabs}
\usepackage{mathrsfs}
\usepackage{bbm}
\usepackage{graphicx}
\usepackage{multirow}
\usepackage{xcolor}

\usepackage{xcolor}
\usepackage{times}
\usepackage{latexsym}
\usepackage{array}
\usepackage[utf8]{inputenc} 
\usepackage[T1]{fontenc}    
\usepackage{url}            
\usepackage{booktabs}       
\usepackage{nicefrac}       
\usepackage{microtype}      
\usepackage{multicol}
\usepackage{multirow}
\usepackage{cancel}
\usepackage{caption}
\usepackage{subcaption}
\usepackage{xspace}
\usepackage{siunitx}
\usepackage{amsfonts}
\usepackage{makecell}

\newcolumntype{L}[1]{>{\raggedright\let\newline\\\arraybackslash\hspace{0pt}}m{#1}}
\newcolumntype{C}[1]{>{\centering\let\newline\\\arraybackslash\hspace{0pt}}m{#1}}
\newcolumntype{R}[1]{>{\raggedleft\let\newline\\\arraybackslash\hspace{0pt}}m{#1}}

\definecolor{Maroon}{cmyk}{0, 0.87, 0.68, 0.32}
\definecolor{lightgrey}{rgb}{0.875, 0.875, 0.875}
\definecolor{darkgreen}{rgb}{0.01, 0.75, 0.24}
\definecolor{lightgreen}{rgb}{0.63, 0.85, 0.61}
\definecolor{lightred}{rgb}{0.95, 0.59, 0.61}
\definecolor{csplit}{rgb}{0.45, 0.45, 0.45}

\definecolor{clabelbox}{rgb}{0.25, 0.25, 0.25}

\usepackage{amssymb}
\usepackage{fontawesome}

\definecolor{caddback}{rgb}{0.90, 0.98, 0.96}
\definecolor{cadd}{rgb}{0, 0.47, 0.34}
\definecolor{cdelback}{rgb}{1, 0.94, 0.92}
\definecolor{cdel}{rgb}{0.83, 0.32, 0.16}

\newcommand{\marco}[1]{}

\def\tabprespace{\vskip -2.4mm}
\def\tabpostspace{\vskip -2.1mm}

\usepackage{xspace}
\usepackage{txfonts}

\definecolor{cexample}{rgb}{0.23, 0.30, 0.45}

\definecolor{ctemplate}{rgb}{0.23, 0.30, 0.45}
\definecolor{cword}{rgb}{0, 0, 0.7}

\newcommand{\gap}{\vspace{0.2mm}}
\newcommand{\examplePairShort}[3]{
    \gap
   \textcolor{csplit}{\{}\ \text{#1}\ \textbf{\textcolor{csplit}{|}}\ \text{#2}\ \textcolor{csplit}{\}}{#3}
    \gap
}

\newcommand{\modelabbrevname}[1]{{{{HERALD}}}{#1}}

\title{
\modelabbrevname: An Annotation Efficient Method to Detect User Disengagement in Social Conversations
}

\author{Weixin Liang$^{1}$ \\
  Stanford University \\
  \texttt{wxliang@stanford.edu} \\\And
  Kai-Hui Liang$^{1}$ \\
  Columbia University \\
  \texttt{kaihui.liang@columbia.edu} \\\And
  Zhou Yu \\
  Columbia University \\
  \texttt{zy2461@columbia.edu} \\
  }

\date{}

\begin{document}
\maketitle
\begin{abstract}
\footnotetext[1]{Equal Contribution. }

Open-domain dialog systems have a user-centric goal: to provide humans with an engaging conversation experience. User engagement is one of the most important metrics for evaluating open-domain dialog systems, and could also be used as real-time feedback to benefit dialog policy learning. Existing work on detecting user disengagement typically requires hand-labeling many dialog samples. We propose HERALD, an efficient annotation framework that reframes the training data annotation process as a denoising problem. Specifically, instead of manually labeling training samples, we first use a set of labeling heuristics to label training samples automatically. We then denoise the weakly labeled data using the Shapley algorithm. Finally, we use the denoised data to train a user engagement detector. Our experiments show that HERALD improves annotation efficiency significantly and achieves 86\% user disengagement detection accuracy in two dialog corpora. 
Our implementation is available at  \url{https://github.com/Weixin-Liang/HERALD/}.

\end{abstract}

\section{Introduction}

Evaluation metrics heavily influence a field's research direction. The ultimate goal of open-domain dialog systems is to provide an enjoyable experience to users. Previous research mainly focuses on optimizing automatic dialog evaluation metrics such as BLEU, which models the distance between the system responses and a limited number of references available. However, it has been shown that these metrics correlate poorly with human judgments~\cite{howNotEval}.

Open-domain dialog system evaluation has long been one of the most difficult challenges in the dialog community for several reasons: (1) The goal of dialog evaluation should be to evaluate users' conversational experience. Existing automatic evaluation metrics such as BLEU are mostly constrained to a static corpus, and do not capture the user experience in a realistic interactive setting. (2) Currently, self-reported user ratings are widely used to evaluate open-domain dialogs. However, self-reported ratings suffer from bias and variance among different users~\cite{liang-etal-2020-beyond}. Although we could tell which dialog system is better by running statistical tests on a large number of noisy ratings, it is challenging to locate dialogs with bad performance reliably. Only by identifying these bad dialogs effectively can we correct errors in these samples to improve dialog system quality.

User engagement has been recognized as one of the essential metrics for open-domain dialog evaluation~\cite{alexa}. Previous research also confirms that incorporating user engagement as real-time feedback benefits dialog policy learning~\cite{DBLP:conf/sigdial/YuNBR16}. One of the most costly bottlenecks of learning to detect user disengagement is to annotate many turn-level user engagement labels~\cite{NanyunPeng}. In addition, the data annotation process becomes more expensive and challenging for privacy-sensitive dialog corpora, due to the privacy concerns in crowdsourcing~\cite{privacyCrowdsourcing}.

To improve annotation efficiency, we reframe the training data annotation process as a denoising problem.  Specifically, instead of manually labeling each training datum, we automatically label the training samples with a set of labeling heuristics. The heuristic functions primarily consist of regular expressions (Regexes) and incorporate open-sourced natural language understanding (NLU) services. Since the automatically generated labels might contain noise, we then denoise the labeled data using the Shapley algorithm~\cite{boxinShapley, boxinShapley2}. We use the Shapley algorithm to quantify the contribution of each training datum, so that we can identify the noisy data points with negative contribution and then correct their labels. Our experiments show that \modelabbrevname{} achieves 86\% accuracy in user disengagement detection in two dialog corpora.

Our proposed framework \modelabbrevname{} is conceptually simple and suitable for a wide range of application scenarios: First, since our model could detect user engagement in real-time (i.e., after each user utterance), our model could be plugged into existing dialog systems as a real-time user experience monitor module. In this way, dialog systems could detect and react to user's disengagement in both open-domain dialogs~\cite{DBLP:conf/sigdial/YuNBR16} and task-oriented dialogs~\cite{DBLP:conf/iwsds/YuRLS17}. During training, our model could also be used as real-time feedback to benefit dialog policy learning~\cite{AlexaEvalDilek}. Second, \modelabbrevname{} could quantify user engagement and be used as an automatic dialog evaluation metric. It could locate dialogs with poor user experience reliably to improve dialog system quality~\cite{NanyunPeng,DBLP:conf/cikm/ChoiAA19}. Third, user engagement is an essential objective of dialog systems, but few dialog datasets with user engagement ratings are available. Our heuristic functions, combined with the proposed workflow, can be readily deployed to annotate new dialog datasets. 

\section{Related Work}

\subsection{Open-Domain Dialog System Evaluation} 
Open-domain dialog system evaluation is a long-lasting challenge. It has been shown that existing automatic dialog evaluation metrics correlate poorly with human judgments~\cite{howNotEval,ademACL,whyWeNeedNewEval}. A well-known reason is that these automatic dialog evaluation metrics rely on modeling the distance between the generated response and a limited number of references available. The fundamental gap between the open-ended nature of the conversations and the limited references~\cite{multiRefTianchengZhao} is not addressed in methods that are lexical-level based~\cite{bleu,rouge,meteor}, embedding based~\cite{greedyMatching,forgues2014bootstrapping}, perplexity based~\cite{menna}, or learning based~\cite{ruber,ademACL}. \citet{UnsupervisedEvaluaitonDialogGPT} simulate user response using DialogGPT and evaluate the probability of user complaint. Given the limitations above, self-reported user ratings are widely used to evaluate open-domain dialogs. However, self-reported ratings suffer from bias and variance among different users~\cite{alexaEval}. Denoising human ratings is still an open research problem~\cite{liang-etal-2020-beyond,ACUTEEVAL}.

\subsection{User Engagement in Dialogs}  
User engagement is commonly defined as the user's willingness to continue conversing with the dialog system~\cite{DBLP:conf/sigdial/YuNBR16,DBLP:conf/iwsds/YuRLS17}. Existing work on measuring user engagement primarily resorts to human rating~\cite{AlexaEvalDilek,DBLP:conf/acl/HancockBMW19}, or proxy metrics. Example proxy metrics include conversation length like number of dialog turns~\cite{alexaEval,alexa}, and conversational breadth like topical diversity~\cite{AlexaTopicEval}. Sporadic attempts have been made to detecting user disengagement in dialogs~\cite{DBLP:conf/interspeech/YuAW04,NanyunPeng,DBLP:conf/cikm/ChoiAA19}. A major bottleneck of these methods is that they require hand-labeling many dialog samples for individual datasets. Although \citet{liang-etal-2020-beyond} denoise user self-reported ratings with the Shapley algorithm for dialog system evaluation, their method cannot be directly applied to dialogs without user ratings as in our setting. Our work is focusing on the problem that it is expensive and difficult to obtain user ratings. The core insight of our work is to reframe the training data annotation process as a process of denoising labels created by heuristic functions pre-defined. To the best of our knowledge, we are the first to combine automatic data labeling with the Shapley algorithm to perform dialog evaluation. Our method could potentially generalize to other classification tasks if different weak labelers are provided.

\begin{figure*}[htb]
\centering
\includegraphics[width=\textwidth]{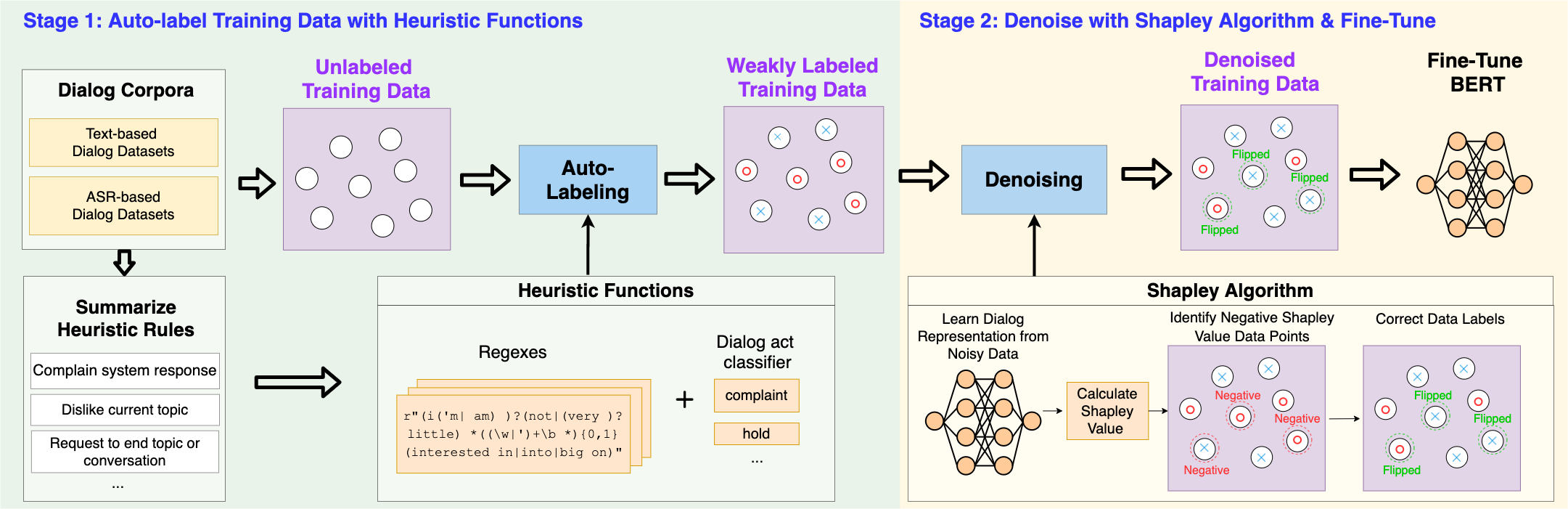} 
    \caption{Schematic of the \modelabbrevname{} two-stage workflow. 
    \textbf{Stage 1:} Auto-label training data with Heuristic Functions. 
    We first design heuristics rules for detecting user disengagement by investigating multiple dialog corpora. 
    The heuristics rules are implemented as heuristic functions based on regular expressions and dialog acts. 
    Then, we use the heuristic function to label the training set automatically. 
    \textbf{Stage 2:} Denoise weakly-labeled training data with Shapley Algorithm. 
    We calculate the Shapley value for each data point and correct the noisy data points with negative Shapely values by flipping their labels. 
    Finally, we fine-tune the model on the denoised training data. 
    } 
    \label{fig:main}
\end{figure*}

\subsection{Learning from Weak Supervision}
Learning from weak supervision reduces annotation costs by utilizing noisy but cost-efficient labels~\cite{snorkel,dataProgramming,liang-etal-2020-beyond}. One of the most popular forms of weak supervision is distant supervision, in which the records of an external knowledge base are heuristically aligned with data points to produce noisy labels for relationship extraction tasks~\cite{DBLP:conf/acl/BunescuM07,DBLP:conf/acl/MintzBSJ09,DBLP:conf/acl/LiangRHVWB18}. Other applications of weak supervision to scene graph prediction~\cite{DBLP:conf/iccv/KrishnaCVBR019}, intent classification~\cite{DBLP:conf/aaai/MallinarSUGGHLZ19}, and medical imaging~\cite{DBLP:conf/nips/VarmaHBKBRR17} have observed similar benefits in annotation efficiency. 
Unlike the existing work, we leverage weak supervision to improve annotation efficiency for detecting user disengagement in social conversations.

\section{Problem Formulation} 
We defined engagement as the degree to which users are willing to continue conversing with the dialog system~\citet{DBLP:conf/sigdial/YuNBR16,DBLP:conf/iwsds/YuRLS17}. We focus on identifying the dialog turns with ``disengaged'' user response, since they usually indicate poor conversation experience. We formulate the user engagement prediction as a binary classification problem: Our goal is to learn a parameterized user engagement predictor $M_\theta$ that, given a dialog turn (along with its dialog context) $x \in \mathcal{X}$, predicts the turn-level user engagement label $y \in \mathcal{Y}=\{0,1\}$, where label $y=1$ means ``disengaged'' and $y=0$ means ``engaged''. We start from an unlabeled train set $ D_\text{train} = \{x_i\}^{N_\text{train}}_1 $ without any label $y_i$. The test set $D_\text{test} = \{(x_i, y_i)\}^{N_\text{test}}_1$ contains the ground-truth label $y_i$. The development set $D_\text{dev}$ has a similar structure as the test set $D_\text{test}$ but the development set can be much smaller than a train set (i.e., $N_\text{dev} \ll N_\text{train}$), making it economical to obtain. Following the general architecture of neural classifiers, we formulate our model $M_\theta = M(\phi, f) = f(\phi(x))$: Here BERT~\cite{bert}-based $\phi$ is a text encoder that maps each dialog turn $x$ to a feature space $\phi(x) \in \mathbb{R}^{d}$. $f$ is the final linear layer with softmax activation.

\begin{table*}[tb]
\tabprespace{}
  \small
  \centering
  \resizebox{\linewidth}{!}{
  \setlength{\tabcolsep}{3pt}
        \begin{tabular}{llrrl}
    \toprule
    \multicolumn{2}{c}{\bf Labeling Heuristics }  &
    \multicolumn{2}{c}{\bf Coverage (\%)}  &
    \multicolumn{1}{c}{\multirow{2}{*}{\bf Example Disengaged User Responses}}\\
    \cmidrule(lr){1-2}
    \cmidrule(lr){3-4}
    \multicolumn{1}{c}{Heuristics Group} &
    \multicolumn{1}{c}{Disengaged intents} & 
    \multicolumn{1}{c}{Gunrock} &
    \multicolumn{1}{c}{ConvAI2} & 
    \\
    \midrule
    \multirow{6}{*}{\begin{tabular}[c]{@{}l@{}} \bf (1) Complain \\ \bf system responses\end{tabular}}
& Complain system repetition & \multirow{6}{*}{1.93} & \multirow{6}{*}{1.95}  & \examplePairShort{You already asked me that.}{I already told you. Remember?}{}  \\
& Complain system ignoring them &   & & \examplePairShort{You're not listening.}{You didn't answer my question.}{}  \\
& Complain system misunderstanding &   & & \textcolor{csplit}{\{} I never said I don't eat my favorite seafood. \textcolor{csplit}{\}} \\
& Not understanding system &   & & \textcolor{csplit}{\{} What are you talking about? \textcolor{csplit}{\}} \\
& Curse system &   & & \textcolor{csplit}{\{} You're dumb. \textcolor{csplit}{\}} \\
& Express frustration &   & & \textcolor{csplit}{\{} Sigh. \textcolor{csplit}{\}} \\
\midrule 
\multirow{2}{*}{
\begin{tabular}[c]{@{}l@{}} \bf (2) Dislike \\ \bf current topic \end{tabular}}
& Express negative opinion & \multirow{2}{*}{1.90} & \multirow{2}{*}{3.45} & \examplePairShort{I don't like music.}{It's boring.}{}  \\
& Show low interests &   & & \textcolor{csplit}{\{} I don't care. \textcolor{csplit}{\}} \\
\midrule 
\multirow{2}{*}{
\begin{tabular}[c]{@{}l@{}} \bf (3) Request to end \\ \bf topic or conversation \end{tabular}}
& Request topic change & \multirow{2}{*}{5.20} & \multirow{2}{*}{2.92} & \textcolor{csplit}{\{} Let's talk about something else. \textcolor{csplit}{\}}  \\
& Request termination &   & & \examplePairShort{Stop.}{Bye.}{} \\
\midrule 
\multirow{4}{*}{
\begin{tabular}[c]{@{}l@{}} \bf (4) End with \\ \bf non-positive responses \end{tabular}}
& End with negative answer & \multirow{4}{*}{20.13} & \multirow{4}{*}{4.86} & \examplePairShort{No.}{I have not.}{}  \\
& End with unsure answer &   & & 
\textcolor{csplit}{\{}\ 
I don't know. 
\textbf{\textcolor{csplit}{|}}\ 
I don't remember. 
\textbf{\textcolor{csplit}{|}}\ 
Well, maybe. 
\textcolor{csplit}{\}} \\
& End with back-channeling &   & & \examplePairShort{Yeah.}{Okay.}{} \\
& End with hesitation &   & & \examplePairShort{Hmm...}{That's a hard one, let me think.}{} \\

    \bottomrule
  \end{tabular}
  }
  \caption{
  Our labeling heuristics designed to capture user disengagement in dialogs. A dialog turn is considered disengaged if any of the heuristic rules apply to the user responses. 
  } 
  \label{table:heuristicRules}
\tabpostspace{}
\end{table*}

\section{Data}
To ensure our framework is generalized to various corpora, we investigate multiple open-domain dialog datasets ranging from ASR-based  (Gunrock~\cite{liang2020gunrock}) to text-based (ConvAI2~\cite{dinan2019second}, Blender~\cite{blender}, and Meena~\cite{menna}) dialog systems. 

\paragraph{Gunrock Movie Dataset}
Gunrock Movie dataset consists of dialog data collected from Gunrock, an ASR-based open-domain social chatbot originally designed for Amazon Alexa Prize~\cite{liang2020gunrock}. The Gunrock dataset comes from a user study where in-lab users were recruited to carry on conversations. We have consent to use the data and we also removed any sensitive information in the conversation. Two dialog experts (co-authors of this paper) randomly annotated 134 dialogs and split them evenly into the test set and development set. In total, the experts labeled 519 turn-level disengaging user responses and 2,312 engaging user responses. They reached a high inter-annotator agreement score~\cite{cohen1968weightedkappa} with kappa $ \kappa = 0.78 $. The training set contains 276 unlabeled dialogs, with 5644 dialog turns. In addition, we ensure that the data annotation is independent of the labeling heuristics collection, so there is no data leakage problem. A full example dialog can be found in Appendix A.4.

\paragraph{ConvAI2 Dataset}
ConvAI2 dataset contains text-based dialog collected from the second Conversational Intelligence (ConvAI) Challenge~\cite{dinan2019second}. We select dialogs from the main eight participated chatbots (Bot 1, 2, 3, 4, 6, 9, 11) and exclude dialogs that are one-sided or shorter than three turns. The dialog experts annotated 207 dialogs in total.  The dialogs are evenly distributed over all the eight bots to ensure system diversity, and are randomly sampled within each bot. The annotated data consist of 209 disengaging turns and 1684 non-disengaging turns. They reached a high inter-annotator agreement score~\cite{cohen1968weightedkappa} with kappa $ \kappa = 0.76 $. We split the annotated dialogs evenly into the test set and develop set. The training set contains 2,226 dialogs, with 18,306 dialog turns.

\paragraph{Google Meena Dataset}
Meena~\cite{menna} is the largest end-to-end neural chatbot so far, trained on $867M$ public domain social media conversations. We study the $93$ example Human-Menna conversations released by Google.

\paragraph{Facebook Blender Dataset}
The Blender bot ~\cite{blender} is an open-domain chatbot with several conversational skills: providing engaging talking points and listening to their partners, displaying knowledge, empathy, and personality appropriately while maintaining a consistent persona. We study the $108$ example Human-Blender conversations released by Facebook.

\section{Method} 
\label{sec:method} 
Our goal is to train a user engagement detector with minimum data annotation efforts. Traditional supervised learning paradigms require annotating many training samples. In addition, it requires additional data annotation to extend the model to a new dialog corpus. To reduce annotation work, we propose HERALD, a two-stage pipeline that annotates large-scale training data efficiently and accurately (Figure~\ref{fig:main}). Instead of hand-labeling training data points, we use heuristic functions to label each training datum automatically. 
The heuristic functions are built upon a set of user disengagement heuristics rules. Since the training data are automatically labeled, their labels would be noisy. We then clean the noisy training data with Shapley algorithm~\cite{jamesShapley} to improve the labeling accuracy. The Shapley algorithm denoises training data by identifying data with wrong labels and flip their labels. Finally, as we received clean training data, we use them to fine-tune a BERT-based model and obtain the final user disengagement detection model. 

\subsection{Stage 1: Auto-label Training Data with Heuristic Functions}
\label{subsec:auto-label}

Since labeling large-scale training data is time-consuming, we propose heuristic labeling functions to label training data automatically. The heuristic functions focus on detecting disengagement from user responses, as it directly indicates poor user experience. To build the heuristics functions, we first summarize the heuristic rules shared among users. We investigate the disengaged dialog turns from the four datasets mentioned above and identify four groups of user disengagement patterns: ``complain system responses'', ``dislike current topics'', ``terminate or change topics'', and ``end with non-positive responses'' (Table \ref{table:heuristicRules}). We then discuss the implementation of heuristics functions.

\subsubsection{Disengagement Heuristic Rules}
\label{subsubsec:disengaged_intents}

\paragraph{Group 1: Complain system responses. } Complaints are an evident sign of user disengagement. We identify six related disengaged intents. The first three intents (``complain system repetition'', ``complain system ignoring them'' and ``complain system misunderstanding'') usually appear when the bot makes errors like repeating the same content, ignoring, forgetting, and misunderstanding the user's response. In these cases, users express their disengagement by indicating the bot's error (e.g. ``You already told me that'', ``You're not listening''). Another intent ``not understanding system'' happens when users cannot understand the system's response (e.g. ``I don't know what you're talking about.''). In the last two intents, users reveal negative emotions by cursing the system (e.g. ``you're dumb'') or express frustration (e.g. ``sigh'') about the conversation.

\paragraph{Group 2: Dislike current topics. } When discussing a given topic, users might show their disengagement by expressing negative opinions or low interest. For example, given the bot's response, ``I write romantic novels under a pen name. '', for users who are not interested in reading, users might say ``reading is boring'', ``I don't like to read'', or ``I'm not interested in this''. We also make sure to handle the corner cases where the user utterance should be labeled as engaged but contains negative opinions. For instance, to respond to the bot's question, ``do you want to not work?'', a user might say, ``Yes. my job is boring. I have to work with mail''. Though the user mentions a negative feeling (``boring''), the user agrees with the bot and shares further information.

\paragraph{Group 3: Terminate or change topics} 
Group 3 considers the cases where users express disengagement to the current topic in a more straightforward fashion. For example, if users are not interested in the current topic, instead of just expressing their dislike to it, they may request to switch topics with ``Let's talk about something else''. In some cases, users might show strong disengagement by requesting to end the conversation if the user is no longer interested in continuing the conversation.

\paragraph{Group 4: End with non-positive responses} A more subtle but common clue of disengagement is when users end the response with non-positive content. For example, non-positive responses like ``I don't know'', ``No'', ``Yeah'', ``uh'', ``Probably'', imply that users do not have much to talk about the current topic. To keep the precision of our heuristics high, we carefully consider the counterexamples. One case is that the user follows up with more responses such as questions (e.g., Bot: ``Have you seen any movies lately? '', User: ``No. Have you?''), and opinion (e.g. Bot: ``What's your favorite animation movie?'', User: ``I don't know, but it might actually be frozen two. My sister loves it.'') in the same dialog turn. These turns should not be labeled as disengaged since the user is still interested in sharing more content or asking follow-up questions. Therefore, we take a conservative approach: we label the dialog turn as disengaged only if no more responses follow the non-positive response.

\subsubsection{Heuristic Functions Implementation}
\label{subsubsec:heuristic_functions}
Next, we discuss how to use heuristic functions to auto-label disengaged user utterances. First, we split user responses into segments since user responses may consist of multiple units with different semantic meanings. We use NLTK Sentence Tokenizer for text-based system, and a segmentation model \cite{chen2018gunrock} for ASR~(Automatic Speech Recognition)-based system as the segmentation tool. We then apply the heuristic functions on each segment to detect disengaged intents. For heuristic groups 1 to 3, if any segment contains a disengaged intent, the user response is auto-labeled as disengaged. For heuristic group 4 (``End with non-positive responses''), we assign disengaged labels only if the disengaged intents are detected in the last segment.

We detect disengaged intents with Regexes. The benefit of using Regexes is that they have minimum dependencies and are easy to modify. We design Regexes for each intent. Following common Regexes complexity metrics~\cite{luo2018marrying}, our Regexes for each intent contains 43.9 Regexes groups and 87.7 \textit{or} clauses on average.

Our framework also supports incorporating additional resources to improve the intent detection accuracy for automatic training data labeling. For example, we can enhance the recall of Regexes intent detection by incorporating existing deep learning-based NLU (Natural Language Understanding) models. Specifically, we re-purpose an open-sourced dialog act classification model~\cite{yu2021midas} to enhance disengagement intent detection: we select 6 out of the 23 supported dialog act labels that are associated with disengaged intents, and map each selected dialog act label to the heuristic groups. The dialog act ``complaint'' is mapped to the heuristic group ``complain system repetition'';``closing'' is mapped to the disengaged intent ``request termination''; ``hold'' to ``hesitation'';``other\_answers'' to ``unsure answer''; ``back-channeling'' to ``back-channeling'', and ``neg\_answer`` to `negative answer`''. If a user utterance is detected with disengaged intent by either Regexes or the deep learning model, then the utterance is auto-labeled as disengaged.

\subsection{Stage 2: Denoise with Shapley Algorithm \& Fine-tune} 
\label{subsec:shapley} 

\paragraph{Overview} 
Next, we denoise the labeled data using Shapley algorithm~\cite{jamesShapley}. Shapley algorithm has been studied in the cooperative game theory~\cite{dubey1975uniqueness} and economics~\cite{shapleyEconomics} as a fair distribution method. Shapley algorithm computes a Shapley value for each training datum, which quantifies the contribution of each training datum to the prediction and performance of a deep network. Low Shapley value data capture outliers and corruptions. Therefore, we can identify and denoise the incorrectly labeled data by computing their Shapley values and fine-tune the model on the cleaned training set.

\paragraph{Shapley Algorithm} 
Shapley algorithm comes originally from cooperative game theory~\cite{dubey1975uniqueness}. Consider a cooperative game with $n$ players $D = \{ 1,...,n \}$ and a utility function $v: 2^{[n]} \rightarrow  \mathbb{R} $ which assigns a reward to each of $2^n$ subsets of players: $v(S)$ is the reward if the players in subset $S \subseteq D$ cooperate. Shapley value defines a unique scheme to distribute the total gains generated by the coalition of all players $v(D)$ with a set of appealing mathematical properties. In our setting, we can consider $ D_{train} = \{(x_i, y_i)\}^{N_\text{train}}_1 $ as $N_\text{train}$ players. We define the utility function $v(S)$ 
as the performance on the development set $D_\text{dev}$. The Shapley value for player $i$ is defined as the average marginal contribution of $\{(x_i, y_i)\}$ to all possible subsets that are formed by other players~\cite{boxinShapley,boxinShapley2}:
    $$ s_{i} = \frac{1}{N} \sum_{S \subseteq D_{train} \setminus \{x_i\} } 
    \frac{1}{\binom{N-1}{|S|}} 
    [v(S \cup \{x_i\}) - v(S)]
    $$

As suggested by the definition of Shapley value, computing Shapley value requires an exponentially large number of computations to enumerate $\mathcal{O}(2^{N_\text{train}})$ possible subsets and train the model $M_\theta$ on each subset, which is intractable. Inspired by~\cite{boxinShapley,boxinShapley2}, \modelabbrevname{} tackles this issue by reducing the deep model $M_\theta$ to a $K$-nearest neighbors ($K$NN) model and then apply the closed-form solution of Shapley value on $K$NN: We reduce our BERT-based classification model $M_\theta = M(\phi, f) = f(\phi(x))$ to a $K$NN by first fine-tuning $M_\theta$ on the auto-labeled training samples. We then use the feature extractor $\phi$ to map each training datum to the feature space $\{\phi(x_i)\}_1^{N_\text{train}}$. We construct a $K$NN classifier in the feature space to compute the closed-form Shapley value.

Next, we discuss the closed-form solution of Shapley value. We first consider a special case where the development set $D_\text{dev}$ only contains one datum $D_\text{dev}=\{ (x_\text{dev}, y_\text{dev}) \}$. Given any nonempty subset $S \subseteq D_\text{train}$, we use the $K$NN classifier to classify $x_\text{dev}$. To do this, we sort the data points in the training set $\{x_i\}_1^{N_\text{train}}$ based on their euclidean distance in the feature space $ \phi(x) $ to the datum in the development set $x_\text{dev}$, yielding 
    $ (x_{\alpha_{1}}, x_{\alpha_{2}}, ..., x_{\alpha_{|S|}} )$
with $x_{\alpha_{1}}, ..., x_{\alpha_{K}}$ as the top-$K$ most similar data points to $x_\text{dev}$. The $K$NN classifier outputs the probability of $x_\text{dev}$ taking the label $y_\text{dev}$ as $P[x_\text{dev} \rightarrow y_\text{dev}]=\frac{1}{K}\sum_{k = 1}^{K} \mathbbm{1}[y_{\alpha_k} = y_\text{dev}]$, where $\alpha_k$ is the index of the $k$th nearest neighbor. We define the utility function as the likelihood of the correct label: 
\begin{align}
\label{eq:utilitySingleDev}
     \nu(S) =\frac{1}{K} \sum_{k=1}^{\min\{K,|S|\}} \mathbbm{1}[y_{\alpha_k(S)} = y_\text{dev}]
\end{align}
\citet{boxinShapley,boxinShapley2} proves that the Shapley value of each training point $s_{\alpha_i}$ can be calculated recursively in $\mathcal{O}(N \log N)$ time as follows: 
\begin{align*}
& s_{\alpha_N} =  \frac{\mathbbm{1}[y_{\alpha_{N}} = y_\text{dev}]}{N}
\\
& s_{\alpha_i} = s_{\alpha_{i+1}}  +  \frac{\min\{K, i\}}{i \times K} \big( \mathbbm{1}[y_{\alpha_i} \!\!= \!\! y_\text{dev}] \!\! - \!\! \mathbbm{1}[y_{\alpha_{i+1}} \!\!= \!\! y_\text{dev}] \big) 
\end{align*}
The above result for a single point in $D_\text{dev}$ could be readily extended to the multiple-point case, in which the utility function is defined by
\begin{align*}
\label{eqn:knn_utility_multiple_dev}
 \nu(S) = \frac{1}{N_\text{dev}}\sum_{j=1}^{N_\text{dev}}\frac{1}{K}\sum_{k=1}^{\min\{K,|S|\}} \mathbbm{1}[y_{\alpha_k^{(j)}(S)} = y_{\text{dev},j}]
\end{align*}
where $\alpha_k^{(j)}(S)$ is the index of the $k$th nearest neighbor in $S$ to $x_{\text{dev},j}$. 
\citet{boxinShapley, boxinShapley2} also prove that the Shapley value in this case is the average of the Shapley value for every single dev point.

\paragraph{Denoising Procedure}
Our denoising procedure works as follows: 
(1) We first fine-tune our BERT-based classification model $M_\theta = M(\phi, f) = f(\phi(x))$ on the auto-labeled training samples. This step injects the knowledge in the labeling heuristic into the model $M_\theta$. 
(2) We then map each auto-labeled training datum to the feature space $\{\phi(x_i)\}_1^{N_\text{train}}$, since we want to apply the closed-form $K$NN formula of Shapley value in the feature space. 
(3) Next, for a binary classification problem, we duplicate each training datum $2$ times with labels $[0,1]$. This generates a large training set $D_\text{large}$ with $2 \times N_\text{train}$ data points, and we note that the origin training set $D_\text{train}$ is a subset of $D_\text{large}$, since $D_\text{large}$ enumerates all $C$ possible labels for each each training datum. 
(4) We then calculate Shapley value for the $2 \times N_\text{train}$ data points in $D_\text{large}$ using the closed-form $K$NN formula. 
(5) We remove the data with negative Shapley value in $D_\text{large}$, and get a cleaned training set $D_\text{clean}$. 
The duplicate-and-remove procedure ``flips'' the labels of the noisy data points with low Shapley value. 
(6) Finally, we fine-tune the classification model $M_\theta$ on $D_\text{clean}$ to get the final user disengagement detection model. 

To sum up, the Shapley value quantifies the contribution of each training datum. Low Shapley value data capture outliers and corruptions that are not consistent with the distribution of other data points. We identify and correct these outliers and corruptions to provide a clean training set.

\section{Experiments}
\label{sec:experiment}

\begin{table}[t]
\footnotesize 
\begin{center}
\tabcolsep=0.03cm
\begin{tabular}{crcccc}
\cmidrule[\heavyrulewidth]{1-6}
\multirow{2}{*}{ \bf No.} 
& \multirow{2}{*}{\bf Method} 
& \multicolumn{2}{c}{\bf Gunrock Movie} 
& \multicolumn{2}{c}{\bf ConvAI2} 
\\
\cmidrule(lr){3-4}
\cmidrule(lr){5-6}
 & & \bf bACC & \bf $F_{2}$Score   
 & \bf bACC & \bf $F_{2}$Score 
 \\
\cmidrule{1-6}
 (1) & Heuristics & 78.32 & 65.09 & 76.58 & 58.16 \\
 (2) & Heuristics (regex only) & 62.81 & 35.46 & 72.04 & 49.90 \\
 (3) & Heuristics (NLU only) & 72.68 & 56.32 & 63.62 & 32.86 \\
\cmidrule{1-6} 
 (4) & Heuristics w/o Group 1 & 78.21 & 64.88 & 71.20 & 48.44 \\
 (5) & Heuristics w/o Group 2 & 77.96 & 64.49 & 75.45 & 56.22 \\
 (6) & Heuristics w/o Group 3 & 71.52 & 55.36 & 71.96 & 49.80 \\
 (7) & Heuristics w/o Group 4 & 58.34 & 23.97 & 68.32 & 42.68 \\
\cmidrule{1-6} 
 (8) & BERT(dev) & 73.98 & 60.74 & 74.97 & 55.40 \\
 (9) & BERT(Auto) & 80.55 & 71.77 & 78.76 & 63.13 \\
 (10) & BERT(Auto+dev) & 80.73 & 72.16 & 80.46 & 64.54 \\
 (11) & \bf \modelabbrevname{} & \bf 86.17\textsuperscript{*} & \bf 80.01\textsuperscript{*} & \bf 86.22\textsuperscript{*} & \bf 70.49\textsuperscript{*} \\
\cmidrule[\heavyrulewidth]{1-6}
\end{tabular}
\caption{
Evaluation results comparison among variants of \modelabbrevname{}. 
* indicates that the model is statistically significantly better than baseline models. 
All numbers in the table are in percentage. 
}
\label{tab:mainresult}
\end{center}
\end{table}

\paragraph{Model Setup} 
We use $K=10$ for the $K$NN Classifier. We use BERT~\cite{bert} as the text encoder $\phi$ of our classification model $M_\theta = M(\phi, f) = f(\phi(x))$. Additional implementation details are included in Appendix.

\paragraph{Model Comparisons and Ablations}
We compare \modelabbrevname{} to its several ablations (Table~\ref{tab:mainresult}) and evaluate the performance on the test set. 
We report balanced accuracy (bACC) and $F_{\beta}$ Score with $\beta=2$~\cite{baeza1999modern}. 
(1) \textit{Heuristics} uses the labeling heuristic function with both Regex and dialog act to predict the test set. 
(2) \textit{Heuristics (Regex only)} uses the labeling heuristic function only with Regex to predict on the test set. 
(3) \textit{Heuristics (NLU only)}  uses the labeling heuristic function only with NLU. 
(4-7) show the ablation of the heuristics function prediction baseline 
by excluding each heuristic group. 
(8) \textit{BERT(dev)} fine-tunes BERT on the expert-annotated development set. 
(9) \textit{BERT(Auto)} fine-tunes BERT on the auto-labeled training samples. 
(10) \textit{BERT(Auto+dev)} fine-tunes BERT on both the auto-labeled training samples and the development set. 
(11) \textit{\modelabbrevname{}} reports the performance of the final model trained on $D_\text{clean}$.

\paragraph{Results} 
Our first takeaway is that our labeling heuristics produce decent predictions and generalize to different datasets. 
As shown in Table~\ref{tab:mainresult}, 
Heuristics prediction (Heuristic, 78.32\%, 76.58\%) is better than the BERT-based model with limited training samples (BERT(dev), 73.98\%, 74.94\%) on both datasets. It also shows that our labeling heuristics are generalizable to different corpora.

Our second takeaway is that learning from a large number of noisy labels works better than learning from a limited number of clean labels. As shown in Table~\ref{tab:mainresult}, BERT fine-tuned on the auto-labeled training set (BERT(Auto), 80.55, 78.76) outperforms BERT fine-tuned on clean but small development set (BERT(dev), 73.98, 74.94) by a large margin. In addition, we also observe that the BERT model fine-tuned on the auto labeled training data (BERT(Auto), 80.55\%, 78.76\%) generalizes beyond the labeling heuristics (Heuristics, 78.32\%, 76.58\%).

Our third takeaway is that using the expert-annotated development set for denoising is more efficient than using the development set as additional training data. After fine-tuning BERT on the weakly labeled training data (BERT(Auto), 80.55\%, 78.76\%), having an additional fine-tuning step using the development set slightly improves the model's performance (BERT(Auto+dev), 80.73\%, 80.46\%). In contrast, using the development set for the Shapley denoising algorithm gives a significant performance gain (\modelabbrevname{}, 86.17\%, 86.22\%).

\begin{figure}[htb]
\includegraphics[width=1.0\linewidth]{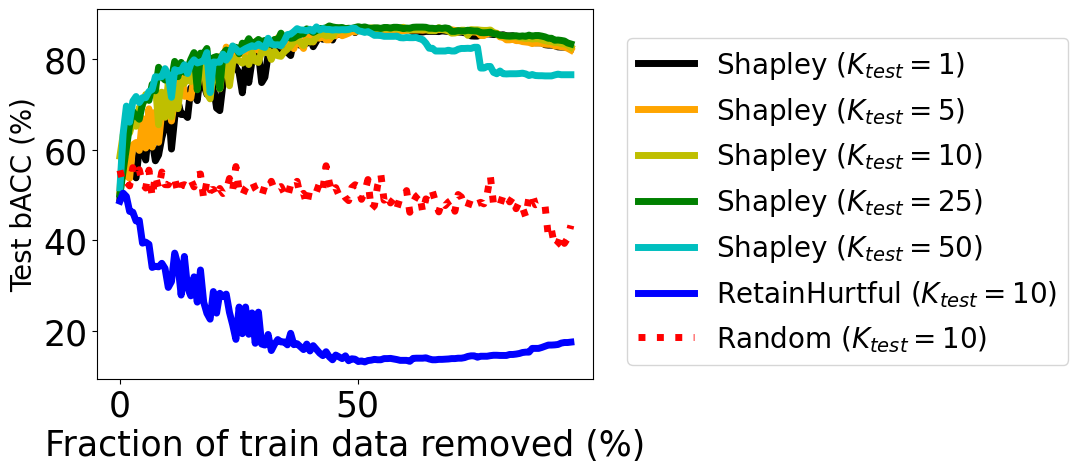}
    \caption{Removing data with low Shapley values (Shapley with $K_{test}$ = 1, 5, 10, 25, 50) improves the performance of the $K$NN in Gunrock Movie Dataset while removing data with high Shapley values and retain data with low Shapley values (``RetainHurtful'') leads to worse performance. } 
    \label{fig:shapleyremove}
\end{figure}

\paragraph{Annotation Cost} 
The cost of annotating the DEV set is small for the Shapley algorithm. For Gunrock Movie Dataset, we used 67 annotated dialogs as the DEV set. For ConvAI2, we used 52 annotated dialogs as the DEV set. The annotation takes less than 1 hour in both cases, which is negligible compared to the cost of annotating all training data.

\paragraph{Heuristics Group Analysis}
We perform ablation studies to analyze the importance of each of the four heuristics groups in Table~\ref{table:heuristicRules}. As shown in Table~\ref{tab:mainresult}, excluding heuristics group 4 leads to the most significant performance drop in both datasets (Heuristics w/o Group 4, 58.34\%, 68.32\%), indicating that ``end with non-positive response'' is the most prevalent form of user disengagement. 

In addition, each heuristics group has different importance in different datasets. For example, dropping heuristics group 1 (``complain system responses'') only leads to a marginal performance drop on the Gunrock Movie dataset but incurs a significant performance drop on the ConvAI2 dataset. We also notice that heuristic group 4 (``End with non-positive responses'') plays a more critical role in the Gunrock Movie dataset than in the ConvAI2 dataset. This might be mainly due to the difference between ASR-based (Gunrock Movie) and text-based (ConvAI2) systems. When asked an open-ended question in ASR-based systems, since users have less time to think, they are more likely to reply with responses such as ``I’m not sure'', ``let me think''. While in text-based systems (ConvAI2), users have more time to think and formulate their responses. Hence, heuristics group 4  covering these responses happen more in Gunrock Movie than ConvAI2.

\paragraph{Generalizability of Heuristic Functions}
The results show that our heuristic functions are generalized to both ASR-based and text-based systems. As indicated in Table \ref{tab:mainresult}, our Regexes reach a decent accuracy of 62.81\% and 72.04\% on the expert annotated test set respectively on Gunrock Movie and ConvAI2 dataset, and thus can serve as a relatively reliable source for auto-labeling.  In addition, although the dialog act model (MIDAS) is initially designed for ASR-based systems and thus has a better performance on the Gunrock Movie data, it should be generalizable to other ASR-based systems, as the six selected dialog acts are general and independent of topics. Therefore, the combination of dialog acts and Regexes should be sufficient to be applied to various corpora.


\paragraph{Shapley Value Analysis}

\begin{figure}[ht!]
\includegraphics[width=1.0\linewidth]{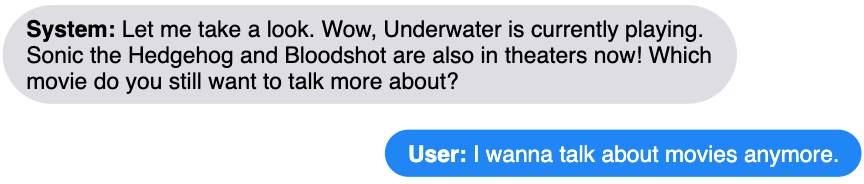}
    \caption{An example dialog turn from the Gunrock Movie dataset with an incorrect auto label ``non-disengaged'' identified by data Shapley. In this case, the user actually says ``I don't wanna talk about movies anymore,'' but an ASR error happens, and thus the labeling heuristics fail to capture this dialog turn.  
    } 
    \label{fig:turn1}
\end{figure}

\begin{figure}[ht!]
\includegraphics[width=1.0\linewidth]{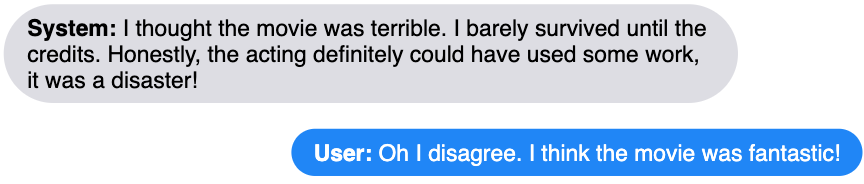}
    \caption{An example dialog turn from Gunrock Movie dataset that is incorrectly auto-labeled as ``disengaged'' because the labeling heuristics see the negative word ``disagree''. This data point is also identified and corrected by data Shapley. 
    } 
    \label{fig:turn2}
\end{figure}

We also present an analysis to show how Shapley denoising works, as shown in Figure~\ref{fig:shapleyremove}. We examine the Shapley value for each training datum in Stage 2. We first show two example dialog turns from the Gunrock Movie dataset with a negative Shapley value in Figure~\ref{fig:turn1} and Figure~\ref{fig:turn2}. In Figure~\ref{fig:turn1}, the dialog turn is incorrectly auto-labeled as ``non-disengaged''. This is because an ASR error happens, and the user utterance ``I don't wanna talk about movies anymore'' is transcribed as ``I wanna talk about movies anymore''.  In Figure~\ref{fig:turn2}, the user says, ``Oh I disagree. I think the movie was fantastic!''. The labeling heuristics see the negative word ``disagree'' and auto-label this turn as ``disengaged''. Both data points are with negative Shapley values and are corrected in Stage 3. 

Next, we present a quantitative analysis of Shapley value. According to the Shapley value, we remove data points one by one, starting from the least valuable (low Shapley values) to the most valuable (high Shapley values). Each time, after removing the data point, we create new $K$NN classifier models on the remaining dialog turns and labels and evaluate them on the test set with expert annotations. As shown in Figure~\ref{fig:shapleyremove}, removing training data with low Shapley values increases the performance to a certain point before convergence for $K$ of all choices. We observe a similar trend when re-training a model on the remaining data. In contrast, removing data randomly or removing data starting from high Shapley values decreases the performance on the test set (``Random'' and ``RetainHurtful'' in Figure~\ref{fig:shapleyremove}). This shows that low Shapley value data effectively capture outliers and corruptions, which further justifies our design choice of denoising with Shapley value.

\paragraph{Alternative Data Valuation Methods} 
We also explored alternative methods to data Shapley like influence function~\cite{influence} and TracIn~\cite{TracIn}: on Gunrock Movie, Influence Functions and TracIn achieve 82.96\% and 83.15\% accuracy, respectively. Both methods outperform BERT(Auto+dev) (80.73\%) significantly but perform slightly worse than HERALD (86.17\%). Overall, results show that our data annotation workflow also works well with other data valuation methods.

\paragraph{Error Analysis}

\begin{figure}[htb]
\includegraphics[width=1.0\linewidth]{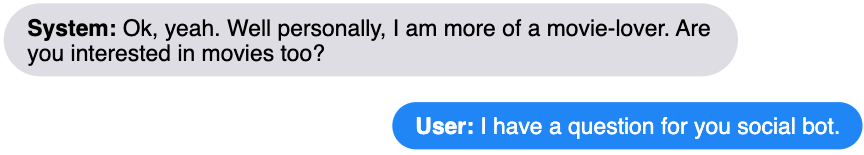}
    \caption{An error case where the low engagement dialog turn that is not captured by \modelabbrevname. 
    } 
    \label{fig:turn3}
\end{figure}

Figure~\ref{fig:turn3} shows an error example of \modelabbrevname, where both the labeling heuristics and the Shapley algorithm fail to identify this turn as low engagement. In this example, the chatbot system asks whether the user is interested in movies, but the user does not directly answer the question. Instead, the user says ``I have a question for you social bot'', indicating that the user does not like the current topic and wants to talk about something else. \modelabbrevname{} fails to identify this dialog turn as low engagement, partly because the Regexes in the ``request topic change'' heuristic rule does not cover this example. One way to fix this error is to upgrade the Regexes. A more general solution is to consider the chatbot system's expectations on user responses conditioned on the chatbot's question. If the chatbot receives an ``unexpected'' user response, then the user is probably not interested in discussing the current topic.

\section{Conclusion}

The ultimate chatbot evaluation metric should be user-centric, as chatbots are there to provide humans with enjoyable experiences. Previously detecting user disengagement typically requires annotating many dialog samples for each individual dataset. We propose a two-stage pipeline \modelabbrevname{} to automatically label and denoise training data and, at the same time, build a user disengagement detector. Our experiment shows that \modelabbrevname{} significantly reduces the annotation cost of a new corpus. \modelabbrevname's disengagement detection results highly correlate with expert judgments on user disengagement in both datasets (86.17\% bACC in Gunrock Movie, 86.22\% in ConvAI2).

\section*{Acknowledgments}
We thank ACL 2021 chairs and reviewers for their review efforts and constructive feedback. We would also like to thank Yu Li and Minh Nguyen for revising the Regexes.

\bibliographystyle{acl_natbib}
\bibliography{ref}

\appendix
\clearpage
\newpage

\section{Appendix}

\subsection{Implementation Details of \modelabbrevname{} }
\label{subsec:implementation}
We use $K=10$ for the $K$NN Regressor. 
We load and fine-tune pre-trained BERT as the feature extractor $\phi$. 
The details of extending BERT to encode multi-turn dialogs are as follows. 
Each dialog turn (along with its dialog context)
is represented as a sequence of tokens 
in the following input format~\cite{DBLP:conf/aaai/LiangTCY20}: 
Starting with a special starting token $[CLS]$, 
we concatenate tokenized user and system utterances in chronological order with $[SEP]$ as the separators for adjacent utterance. 
In other words, we represent each dialog as a sequence: $[CLS]$, $S_{1,1}$, $S_{1,2}$, $...$, $[SEP]$, $U_{1,1}$, $U_{1,2}$, $...$, $[SEP]$, $S_{2,1}$, $S_{2,2}$, $...$, $[SEP]$ 
where $S_{i,j}$ and $U_{i,j}$ are the $j^{th}$ token 
of the system and user utterance in the $i^{th}$ turn. 
Following BERT, 
we also add a learned embedding to 
every token indicating whether it comes from user utterances or system utterances
. 
In addition, 
since the disengaging class 
and the non-disengaging class are imbalanced, 
we up-sample the disengaging dialog turns for both the training set and the development set. 
Though it is also possible to handle the imbalanced classes by adding weights for two classes, 
we did not take this approach because 
we do not have a closed-form solution for calculating the shapley value for weighted $K$NN in $\mathcal{O}(N \log N)$ time. 
Improving the architecture of HERALD and extending HERALD to other machine learning tasks~\cite{liang2021neural,DBLP:conf/emnlp/LiangZY20,DBLP:journals/corr/abs-2011-10731,2021graphvqa} are interesting directions of future work.

\subsection{Reproducibility}
The source code of \modelabbrevname{} can be found in the supplementary materials. We run experiments on a server of eight GTX 1080 GPUs. The average runtime for all stages of \modelabbrevname{} is less than 10 minutes. The number of parameters is similar to BERT. We use the default hyperparameters of BERT. 
The public examples of Google Meena Dataset can be downloaded from \url{https://github.com/google-research/google-research/blob/master/meena/meena.txt} 
The public examples of Facebook Blender Dataset can be downloaded from \url{https://parl.ai/projects/recipes/chatlog_2.7B_render.html} 
The public examples of ConvAI2 Dataset can be downloaded from \url{http://convai.io/data/data_volunteers.json} 
and 
\url{http://convai.io/data/summer_wild_evaluation_dialogs.json}



\paragraph{Additional Shapley Value Analysis}

\begin{figure}[htb]%
    \centering
    \subfloat[\centering 
    Denoising with Shapley Value in Gunrock Movie Dataset
    ]{\includegraphics[width=0.4\textwidth]{./figures/gunrock_shapley.png} }%
    \qquad
    \subfloat[\centering Denoising with Shapley Value in ConvAI2 Dataset]{\includegraphics[width=0.4\textwidth]{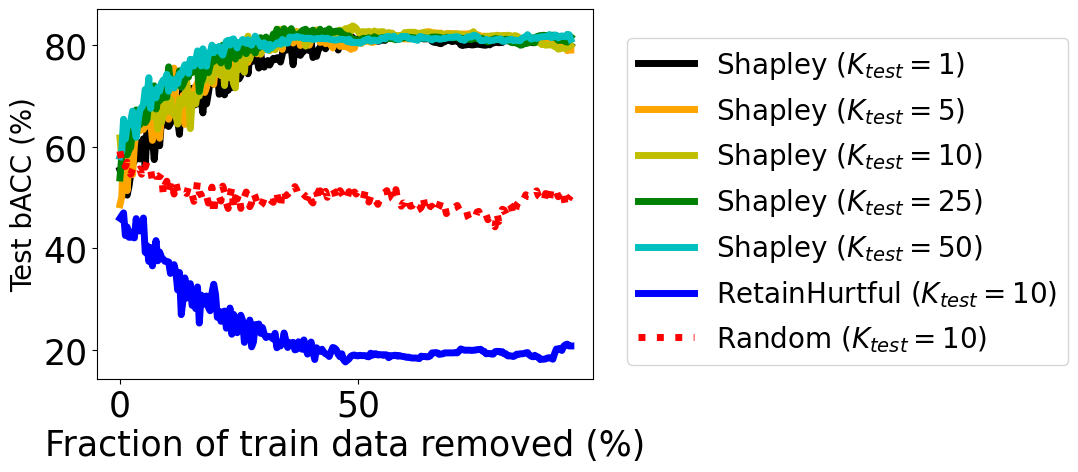} }%
    \caption{Removing data points with low Shapley value improves the performance of the $K$NN classifier. 
    } 
    \label{fig:shapleyremove3}
\end{figure}

We also present addition analysis to show how Shapley denoising works as shown in Figure~\ref{fig:shapleyremove3}. 
We present the experiments on both Gunrock Movie Dataset and ConvAI2 Dataset. 
Figure~\ref{fig:shapleyremove3} presents a quantitative analysis of Shapley value. 
According to the Shapley value, 
we remove data points one by one starting from the least valuable to the most valuable. 
Each time, after the data point is removed, 
we create new $K$NN classifier models on the remaining dialog turns and labels and evaluate them on the test set with expert annotations. 
As shown in Figure~\ref{fig:shapleyremove3},
removing training data with low Shapley values increases the performance to a certain point before convergence for $K$ of all choices. We observe a similar trend when re-training a model on the remaining data. 
In contrast, removing data randomly or removing data from the most to least valuable data decreases the performance on the test set. 
This shows that low Shapley value data effectively capture outliers and corruptions, which further justifies our design choice of denoising with Shapely value.

\subsection{Addition Dialog Examples}
\begin{figure}[htb]%
\centering
\includegraphics[width=1.0\linewidth]{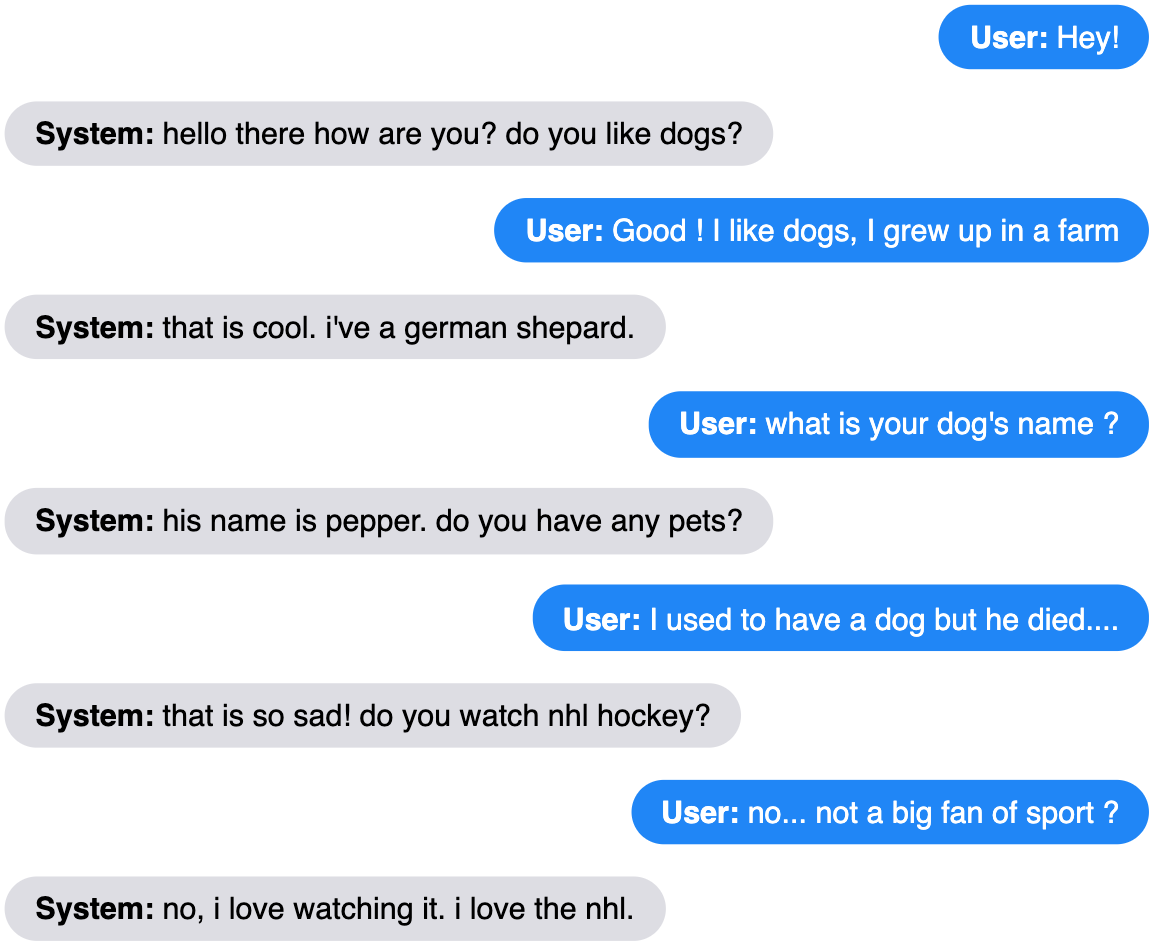}
    \caption{A full example from ConvAI Dataset. 
    } 
    \label{fig:convaifullexample}
\end{figure}

\begin{figure}[htb]%
\centering
\includegraphics[width=0.95\linewidth]{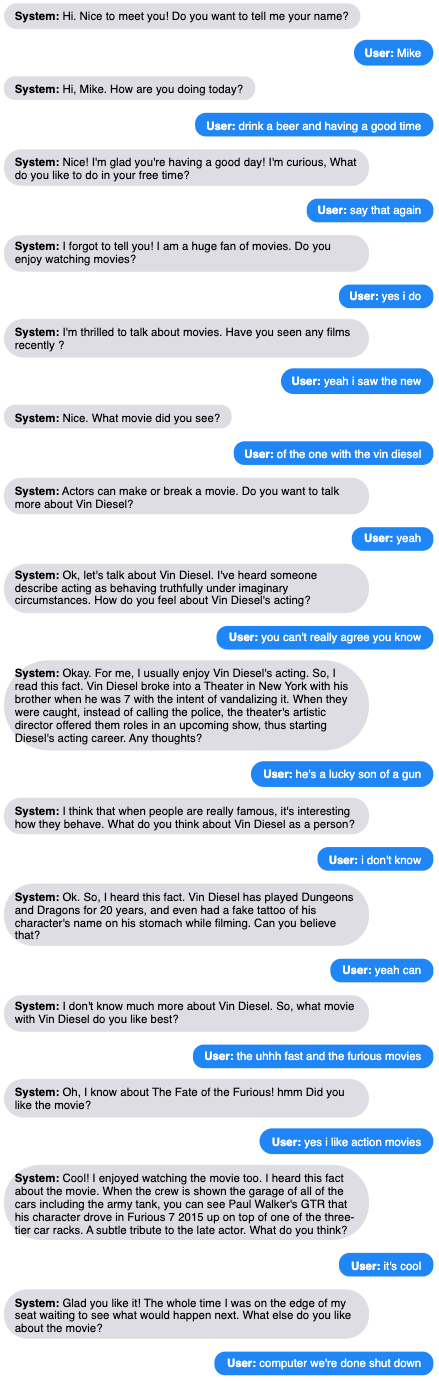}
    \caption{A full example from Gunrock Movie Dataset. 
    } 
    \label{fig:fullexample}
\end{figure}

We show additional dialog examples. 
Figure~\ref{fig:convaifullexample} shows a full dialog example from ConvAI dataset. 
Figure~\ref{fig:fullexample} shows a full dialog example from Gunrock Movie dataset. 

\end{document}